\documentclass{article}

\usepackage{PRIMEarxiv}

\usepackage[utf8]{inputenc} 
\usepackage[T1]{fontenc}    
\usepackage{hyperref}       
\usepackage{url}            
\usepackage{booktabs}       
\usepackage{amsfonts}       
\usepackage{nicefrac}       
\usepackage{microtype}      
\usepackage{lipsum}
\usepackage{fancyhdr}       
\usepackage{graphicx}       
\graphicspath{{media/}}     

\title{Financial Time-Series Forecasting: Towards Synergizing Performance And Interpretability Within a Hybrid Machine Learning Approach
}

\author{
  Shun Liu$^{1\dagger}$\thanks{First Author $^{\dagger}$Corresponding Author} \\
  \texttt{kevinliuleo@gmail.com} \\
  \And
  Kexin Wu$^{5}$\\
  \texttt{kw634@cornell.edu}\\
  \And
  Chufeng Jiang$^{4}$\\
  \texttt{chufeng.jiang@utexas.edu}\\
  \And
  Bin Huang$^{3}$\\
   \texttt{bin@smu.edu}\\
  \And
  Danqing Ma$^{2}$\\
  \texttt{3530761316qq@gmail.com}\\
  $^{1}$Department of Computer Science, Shanghai University of Finance and Economics, Shanghai, China \\
  $^{2}$Department of Computer Science, Stevens of Institute of technology, Hoboken, USA\\
  $^{3}$Electrical and computer engineering, Southern Methodist University, Dallas, USA\\
  $^{4}$Department of Computer Science, The University of Texas at Austin, Texas, USA\\
  $^{5}$Independent Researcher, New York, USA\\
}

\begin{document}
\maketitle

\begin{abstract}
In the realm of cryptocurrency, the prediction of Bitcoin prices has garnered substantial attention due to its potential impact on financial markets and investment strategies. This paper propose a comparative study on hybrid machine learning algorithms and leverage on enhancing model interpretability. Specifically, linear regression(OLS, LASSO), long-short term memory(LSTM), decision tree regressors are introduced. Through the grounded experiments, we observe linear regressor achieves the best performance among candidate models. For the interpretability, we carry out a systematic overview on the preprocessing techniques of time-series statistics, including decomposition, auto-correlational function, exponential triple forecasting, which aim to excavate latent relations and complex patterns appeared in the financial time-series forecasting. We believe this work may derive more attention and inspire more researches in the realm of time-series analysis and its realistic applications.
\end{abstract}

\section{Introduction}

The evolution of machine learning has fundamentally transformed a myriad of fields, showcasing its remarkable versatility and power in tackling complex problems. From traffic sign recognition \cite{b1} and cancer gene data classification \cite{b2}, to the challenges of autonomous navigation at unsignalized intersections \cite{b3} and real-world storm prediction \cite{b4}, the impact of machine learning is profound and far-reaching. These diverse applications demonstrate the capability of machine learning to not only analyze but also predict and interpret vast and complex datasets across various domains.

In transportation, for instance, machine learning models have been instrumental in improving safety and efficiency, as seen in traffic sign recognition systems \cite{b1}. Similarly, in healthcare, deep learning techniques have enabled significant advances in understanding genetic data, thereby enhancing cancer diagnosis and treatment \cite{b2}. The field of autonomous navigation has also benefited greatly, with reinforcement learning and model predictive control approaches playing a crucial role in navigating complex environments \cite{b3}. Moreover, the predictive power of machine learning is exemplified in meteorology, where advanced models are used for accurate storm prediction, aiding in disaster preparedness and response \cite{b4}.

The application of machine learning extends to the realm of urban planning and transportation, where deep sequential models have been used for utility-based route choice behavior modeling \cite{b5}. This illustrates how machine learning can assist in understanding and predicting human behavior in complex urban environments. Furthermore, the field of time-series analysis has seen significant advancements through machine learning, particularly in dealing with nonlinear and non-stationary data, as demonstrated by Huang et al. \cite{b6}. This is particularly relevant in financial markets, where time-series data is abundant and complex.

In this report, we focus on the application of machine learning in the financial sector, particularly in the realm of cryptocurrencies like Bitcoin. The cryptocurrency market, characterized by its high volatility and unpredictable nature, presents a unique challenge for traditional financial models. Accurate prediction of Bitcoin prices is crucial for investors, traders, and regulatory bodies, and has wider implications for financial stability and economic planning.

Our approach is multifaceted: we begin with a comprehensive examination of time-series data preprocessing techniques, which are crucial for accurate analysis and forecasting in financial markets. Drawing parallels from the work of Ching and Phoon \cite{b7}, we emphasize the importance of understanding the underlying patterns in time-series data, such as those related to climate change \cite{b8}, for making accurate predictions. We also explore various machine learning models, including tree-based regressors and neural network architectures, to capture the complex dynamics of Bitcoin price movements. This is akin to the methodologies used in multivariate financial time series forecasting \cite{b9} and hybrid financial time series prediction models \cite{b10}.

Moreover, we discuss the criticality of explainability in our models. As the financial sector relies heavily on understanding and interpreting wdata, our goal is to provide clear, actionable insights alongside accurate predictions. This aligns with the broader movement towards explainable AI in various fields, including energy consumption forecasting \cite{b11} and energy management optimization \cite{b12}.

Our exploration also includes the consideration of external factors impacting the cryptocurrency market, drawing inspiration from diverse fields where external influences are significant, such as call center arrivals forecasting \cite{b13} and energy consumption prediction \cite{b14}. Finally, we highlight the ethical dimensions of our work, recognizing the potential impacts on market stability and investor behavior, a concern also evident in other domains such as real-time resource load prediction \cite{b15}.

Through this comprehensive approach, our research aims to contribute significantly to the field of cryptocurrency analysis, providing a robust framework for Bitcoin price prediction and setting the stage for future research that could extend these methods to other cryptocurrencies and integrate them with traditional financial models. This work serves as a crucial link between the evolving world of digital currencies and the established realm of financial analysis and decision-making.

\label{introduction}



\section{Methodology}
\label{method}
\subsection{EDA and Data Preprocessing}
We gather historical Bitcoin price data from multiple sources and perform data preprocessing to ensure its quality and consistency. The collected data spans from 2013 to 2018 and contains approximately 3,000 trading days. Remind the task of this report is to forecast the closing prices(annotated as dependent variable) based on the past market performance(denoted as independent variables), which is characterized as following:

\begin{itemize}
    \item \textbf{Independent Variables}: date, open prices, high prices, low prices, market capacity, training volume(per day)
    \item \textbf{Dependent Variable}(i.e. target variable): bitcoin closing prices.
\end{itemize}

Specifically, data sampling frequency is 

\begin{figure}
    \centering
    \includegraphics[width=14cm]{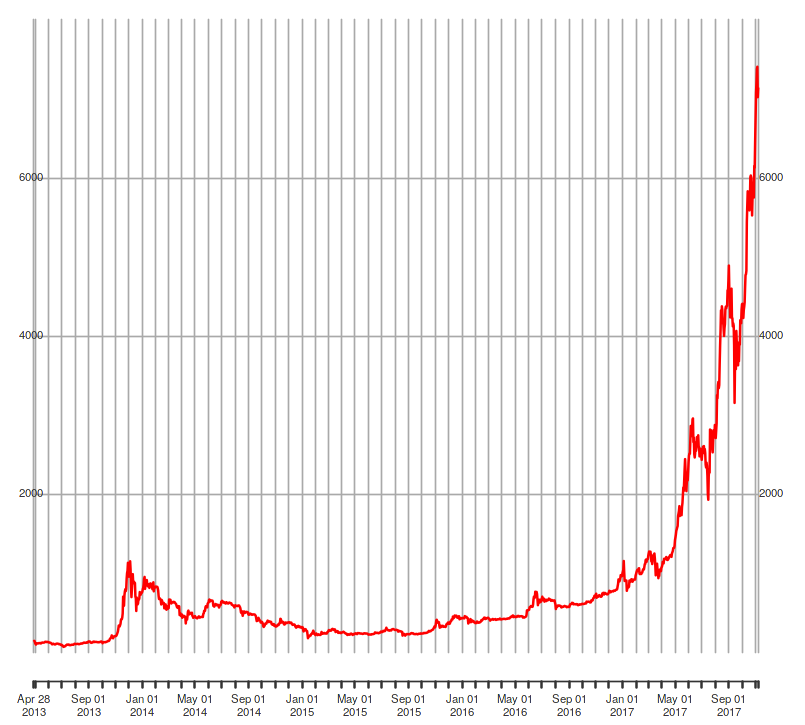}
    \caption{Overview of the closing prices in the collected dataset.}
    \label{fig:enter-label}
\end{figure}

\subsubsection{Decomposition}
The decomposition process allows us to isolate and analyze each component separately, providing insights into the underlying patterns, trends, and irregularities in the data. This information can be useful for forecasting, identifying anomalies, and gaining a deeper understanding of the time series behavior. By visualizing the decomposed components, we can observe the individual patterns and fluctuations of the trend, seasonality, and random components, which helps in interpreting and analyzing the time series data more effectively. After decomposition, we can attain four primary components:

\textbf{Trend} represents the long-term pattern or direction of the time series data. It captures the overall upward or downward movement over an extended period. Trends can be linear (constant slope) or nonlinear (curved or irregular).

\textbf{Seasonality} captures regular patterns that repeat at fixed intervals within the time series. These patterns could be daily, weekly, monthly, quarterly, yearly, or any other consistent pattern. Seasonality can be influenced by factors such as holidays, weather, or cultural events.

\textbf{Random} also known as the residual or error term, represents the unpredictable fluctuations or noise in the time series. It includes all the factors that cannot be explained by the trend or seasonality. Random variations are typically assumed to be independent and identically distributed (i.i.d.) and have a mean of zero.

\textbf{Observed} is the actual time series data that we initially start with. It is a combination of the trend, seasonality, and random components. By decomposing the time series, we separate out these different components to better understand their individual contributions.

\begin{figure}
    \centering
    \includegraphics[width=12cm]{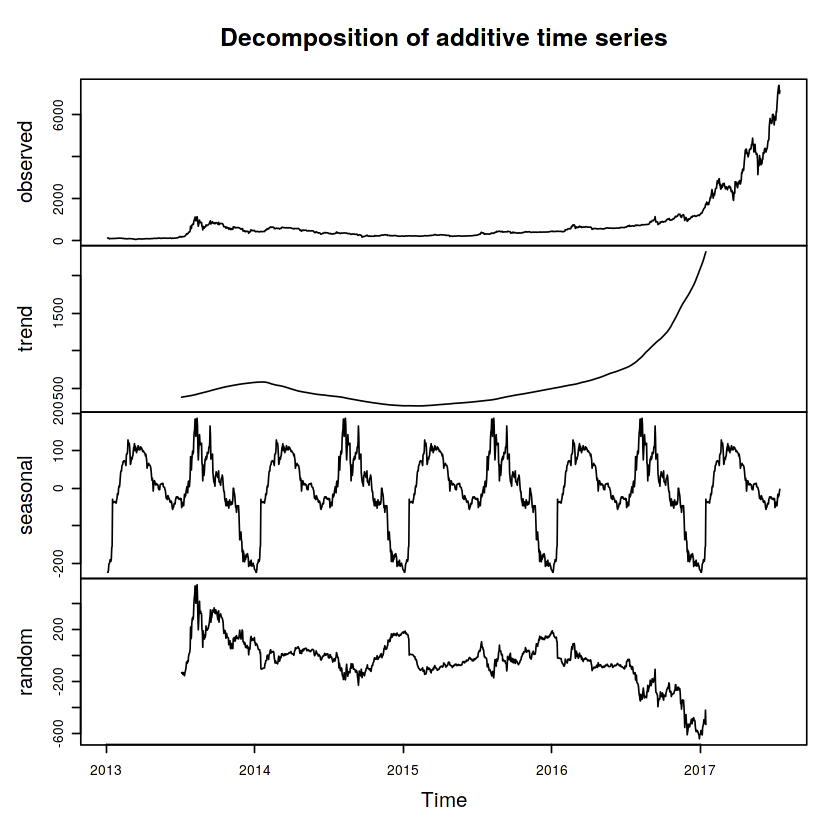}
    \caption{Decomposition of historical prices, which is consisted of four components: 'trend' comopnent deliver a long-term picture of the prices; 'seasonality' identifies the regular pattern for a fixed amount of time range; 'random' component address the existence of noises or fluctuations in the raw dataset, hence introduce some uncertainty to the model, this stage conjecture samples are independent and identically distributed(i.i.d); 'observed' component reflects real marketing data, which integrates the rest three components to formulate a comprehensive depictions.}
    \label{fig:enter-label}
\end{figure}

\subsubsection{Autocorrelation Function(ACF)}
In this part, we use the intuitions from ACF\cite{a2} to study some thoughtful insights. ACF and PACF(Partial Autocorrelation Function) depicts the lag autocorrelation of a time series. They also assume underlying strationarity of time series, which is typically judged by hypothesis checking: if $p-value$ is greateer than a critical value of 0.05, then we claim that the data has a unit root, hence non-stationary; otherwise the data is stationary.

\begin{figure}
    \centering
    \includegraphics[width=8cm]{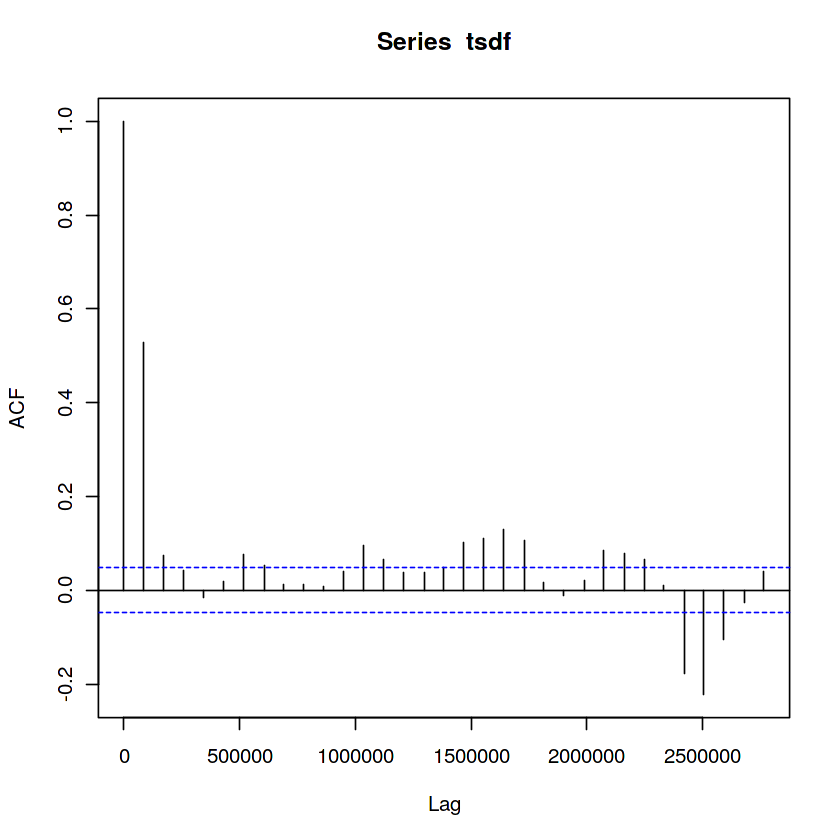}
    \includegraphics[width=8cm]{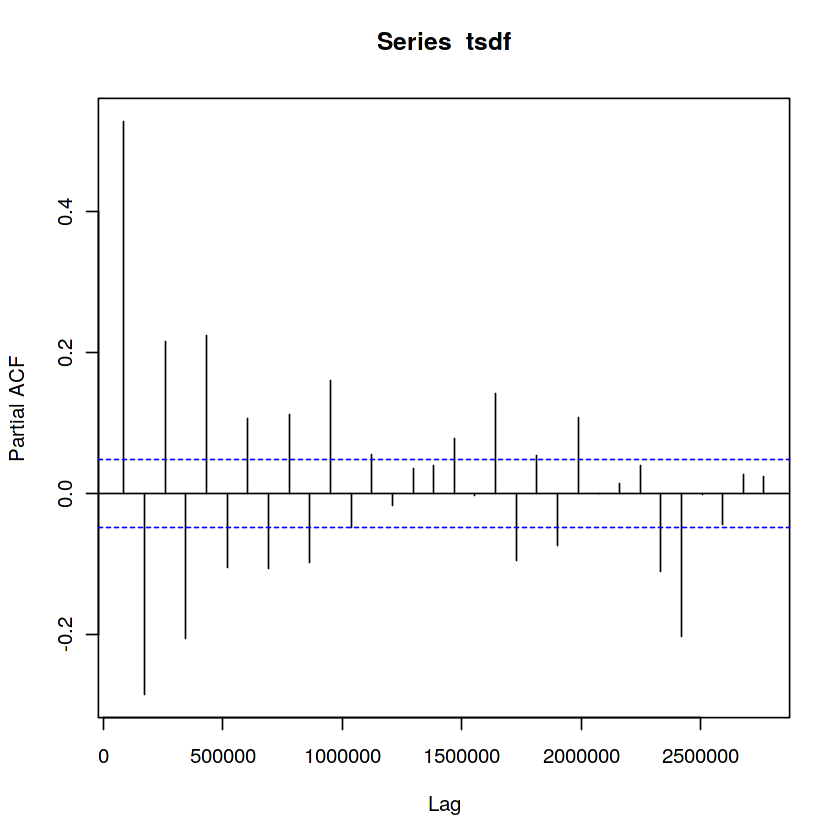}
    \caption{Autocorrelation function(ACF) and partial autocorrelation function(PACF) between observed time series and lagged time series.}
    \label{fig:enter-label}
\end{figure}

\subsection{Time-Series Forecasting}
\textbf{Exponential triple smoothing} is also known as Holt-Winters method, is a popular time series forecasting technique that extends the Holt method by incorporating seasonality. Like simple exponential smoothing\cite{a3,a9,a10,a11} and Holt method, the Holt-Winters method uses smoothing parameters to estimate the level, trend, and seasonal components of a time series.

The Holt-Winters method uses three smoothing parameters: $\alpha$, $\beta$, and $\gamma$. Alpha and beta are similar to the parameters used in the Holt method and control the smoothing of the level and trend components, respectively. Gamma controls the smoothing of the seasonal component and is used to estimate the seasonal pattern in the data.

The model can be expressed as follows:

Level equation: $$L(t) = \alpha * (Y(t) - S(t-m)) + (1 - \alpha) * [L(t-1) + T(t-1)]$$

Trend equation: $$T(t) = \beta * [L(t) - L(t-1)] + (1 - \beta) * T(t-1)$$

Seasonal equation: $$S(t) = \gamma * (Y(t) - L(t)) + (1 - \gamma) * S(t-m)$$

Forecast equation: $$F(t+h) = L(t) + h * T(t) + S(t+h-m+(1-m\%2)\/2)$$

where:
\begin{itemize}
    \item$Y(t)$: the observed value at time $t$
    \item$L(t)$: the estimated level at time $t$
    \item$T(t)$: the estimated trend at time $t$
    \item$S(t)$: the estimated seasonal factor at time t
    \item$\alpha$, $\beta$, and $\gamma$: smoothing parameters between 0 and 1
    \item$m$: the number of seasons in a year (e.g., 12 for monthly data)
    \item$h$: the forecast horizon
\end{itemize}

The seasonal equation introduces a new term to account for periodic fluctuations in the data. The seasonal factor S(t) represents the deviation of Y(t) from the average level at time t, and it is estimated using a moving average of length m. The forecast equation uses the estimated level, trend, and seasonal factors to make predictions for future values.

Like the Holt method, the Holt-Winters method requires setting the smoothing parameters alpha, beta, and gamma. These can be determined by minimizing a measure of forecast error, such as mean squared error or mean absolute percentage error. The optimal values of the parameters can also be chosen using cross-validation techniques.

Overall, the Holt-Winters method is a powerful forecasting technique that can be used to make accurate predictions for time series data with trends and seasonal patterns. By incorporating both the level, trend, and seasonal components of a time series, it can provide reliable forecasts that can be used for a variety of applications.

\section{Experiment}
\subsection{Preprocessing}
\begin{figure}
    \centering
    \includegraphics[width=8cm]{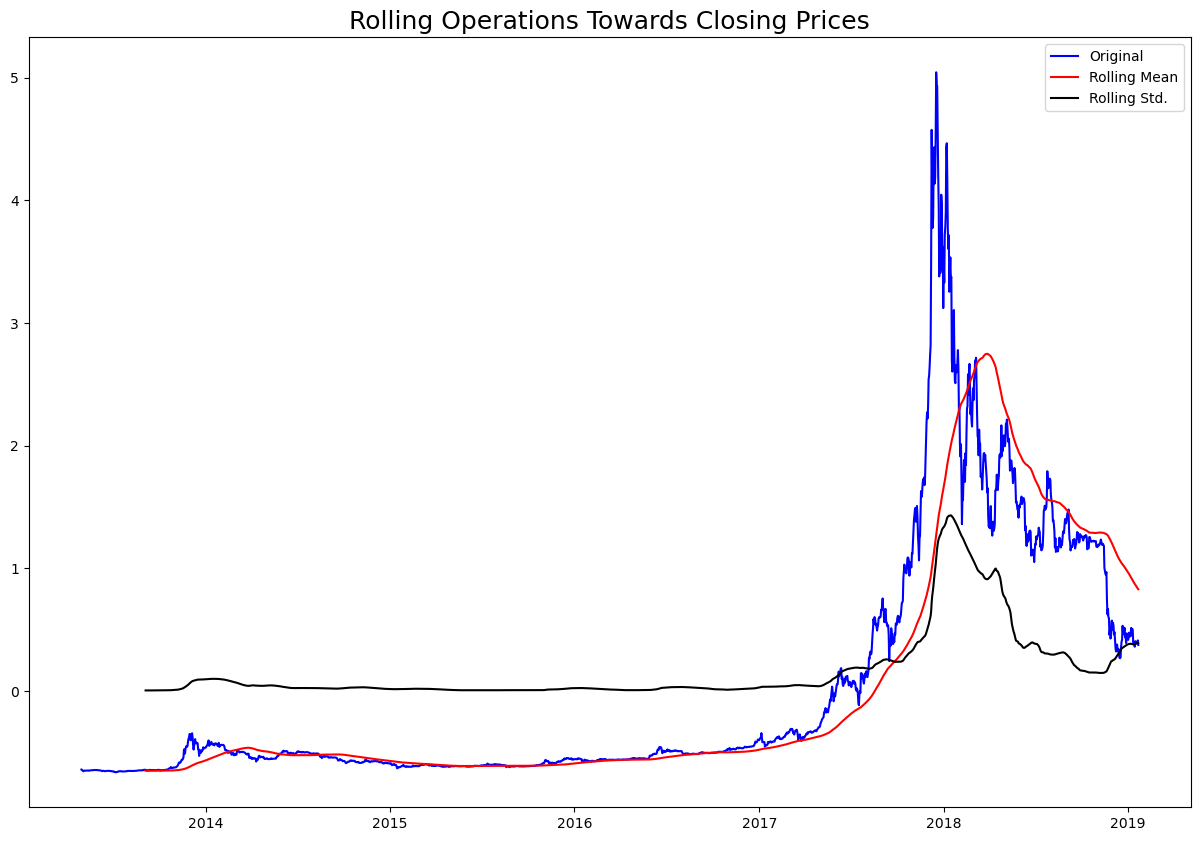}
    \includegraphics[width=8cm]{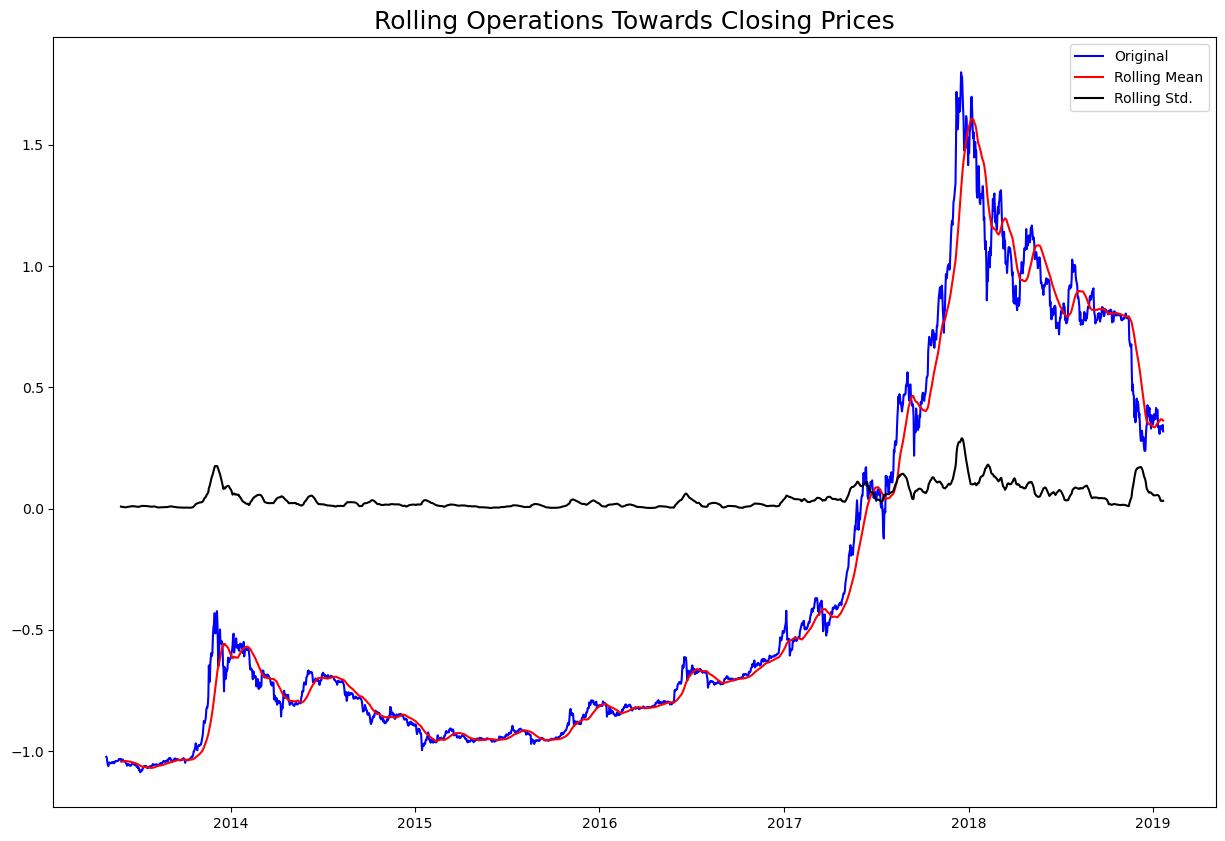}
    \caption{Rolling factors(mean, std.) of bitcoin closing prices in the normal scale(left) and log scale(right).}
    \label{figure-eda-rolling}
\end{figure}

\textbf{Normalization.} To restrict the input into the identical scale, we employ \textit{MinMaxScaler} to conduct normalization respectively. For rolling features, Figure \ref{figure-eda-rolling} reflects the relative fluctuation and periodical average of the bitcoin closing prices. 

\textbf{Data Splitting.} During the grounded experiment, we split the raw data into three subsets for training, testing, and validating separately. Validation set is adopted to simulate the statistics appeared in the real world, test set is utilized to check the model performance. Moreover,  for those MLP-like architecture(CNN, LSTM). For the details, training and test splitting ratio is fixed to 0.2.

\subsection{Protocol and Metrics}
The experimental environment is configured with 2 * RTX 3090Ti for the hardware; The software is configured with torch version 2.0.1+cu118(GPU), python version of 3.9.0. 

Evaluation and Validation: We rigorously evaluate the performance of our proposed approach through cross-validation, backtesting, and comparison with existing models, demonstrating its superiority in forecasting Bitcoin price dynamics.
Case Studies and Real-world Applications: We present case studies and real-world applications to showcase the practical utility and effectiveness of our methodology in supporting investment decisions\cite{a1,a4,a5} and risk management strategies\cite{a4} in the cryptocurrency market.

\subsection{Candidate Models}\label{sec-exp}

\textbf{Linear Regression with Moving Average(MA).}
With linear regression(LR) model, we focus on excavating the temporal linearity of the time-series in 'bitcoin closing prices', which indicates how reliant of the data points to the previous ones. 
Experiment setting and variable definitions are introduced. For dependent variable, to our goal, are defined as the 'bitcoin closing prices apart from the latest 5 days' in order to detect the correlation across time span, shifting window has the size of 5. This means we only use the historic statistics excluding latest 5 days to predict the closing prices; The independent variable is naturally 'current bitcoin closing prices'.

Through the experiment, the \textbf{\textit{linear regression score}} achieves impressive \textbf{0.9998}, which shows the independent variables are strongly correlated with the dependent variable.

There are some concerns on the potential overfitting, to address the issue we employ regularization techniques as well. Specifically, \textbf{\textit{Lasso Regression}} with \textit{L1 regularization}is introduced. With expectation, the mean absolute error drops from 96.20 to 91.80.

\textbf{Decision Tree.} To examin how tree-based regressor adapt to the financial time-series forecasting tasks, decision tree is also proposed with the following configurations(note that the optimal hyperparameter settings are retreived by GridSearch method, which performs parallel experiments on different parameter protocols and end with the best collocation, based on metrics performance):

\begin{itemize}
    \item max\_depth(The maximum depth of the tree): 15
    \item min\_samples\_leaf(The minimum number of samples required to be at a leaf node): 10
    \item min\_impurity\_decrease(A node will be split if this split induces a decrease of the impurity greater than or equal to this value): 0.5
    \item criterion(The function to measure the quality of a split): 'mean\_squared\_error'
    \item splitter(The strategy used to choose the split at each node): 'best'
    \item max\_features(The number of features to consider when looking for the best split): 'auto'
\end{itemize}

\textbf{Long-Short Term Memory(LSTM).} The proposed LSTM model consists of 1 \textit{LSTM} layer, 1 \textit{Dense} layer. Training loss is defined as the mean of squared errors, scoring metrics is fixed to the mean of absolute errors. Note that training optimizer is \textbf{\textit{Adam}}, a complex which integrates the strengths of \textbf{\textit{RMSProp}} and \textbf{AdaGrad}. Through 100 training epoch, the results of LSTM has been shown in Figure \ref{figure-result}. Unluckily, proposed LSTM model fails to converge during the training, we observe that training and validation loss are hard to drop synchronously. The reason behind this phenomenon maybe the inefficiency of data, which cause the underfitting.

\begin{figure}
    \centering
    \includegraphics[width=18cm]{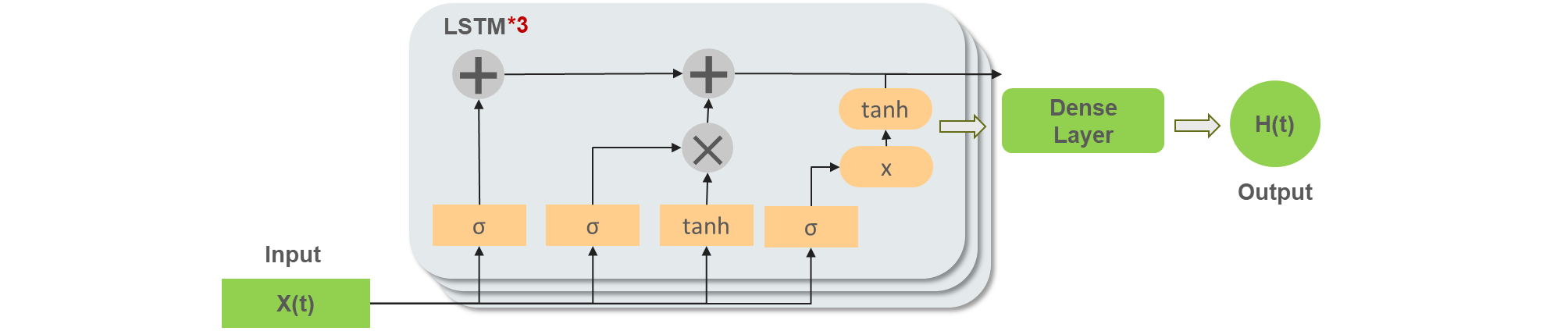}
    \caption{Multi-layer LSTM architecture. Unlike conventional LSTM model, we concatenate more LSTM layers in the single model, and introduce dropout regularization to avoid overfitting, which is essential to the financial time-series forecasting.}
    \label{figure-network}
\end{figure}

\begin{figure}
    \centering
    \includegraphics[width=12cm]{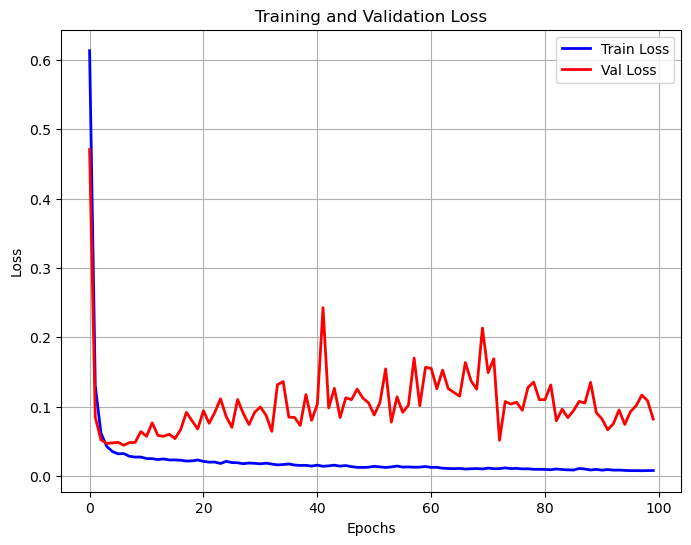}
    \caption{Training and validation loss of LSTM model. From the figure, training loss drops as epochs increasing, but validation loss fluctuates at the tailing epochs, suggest that the model is difficult to converge during training process, therefore we claim that there are issues of data insufficiency or model definitions to address.}
    \label{figure-loss-lstm}
\end{figure}

\subsection{Results}
\begin{figure}
    \centering
    \includegraphics[width=12cm]{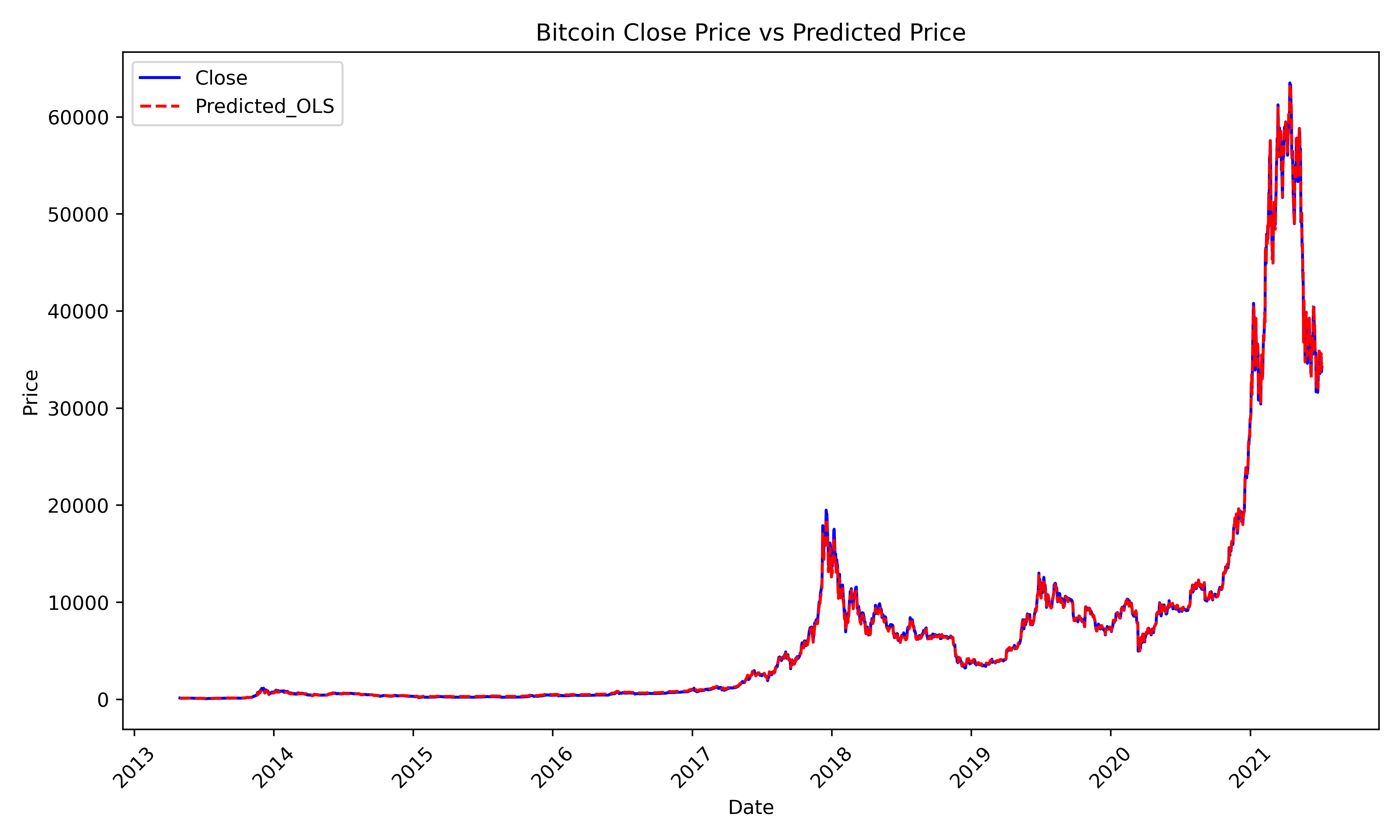}
    \includegraphics[width=12cm]{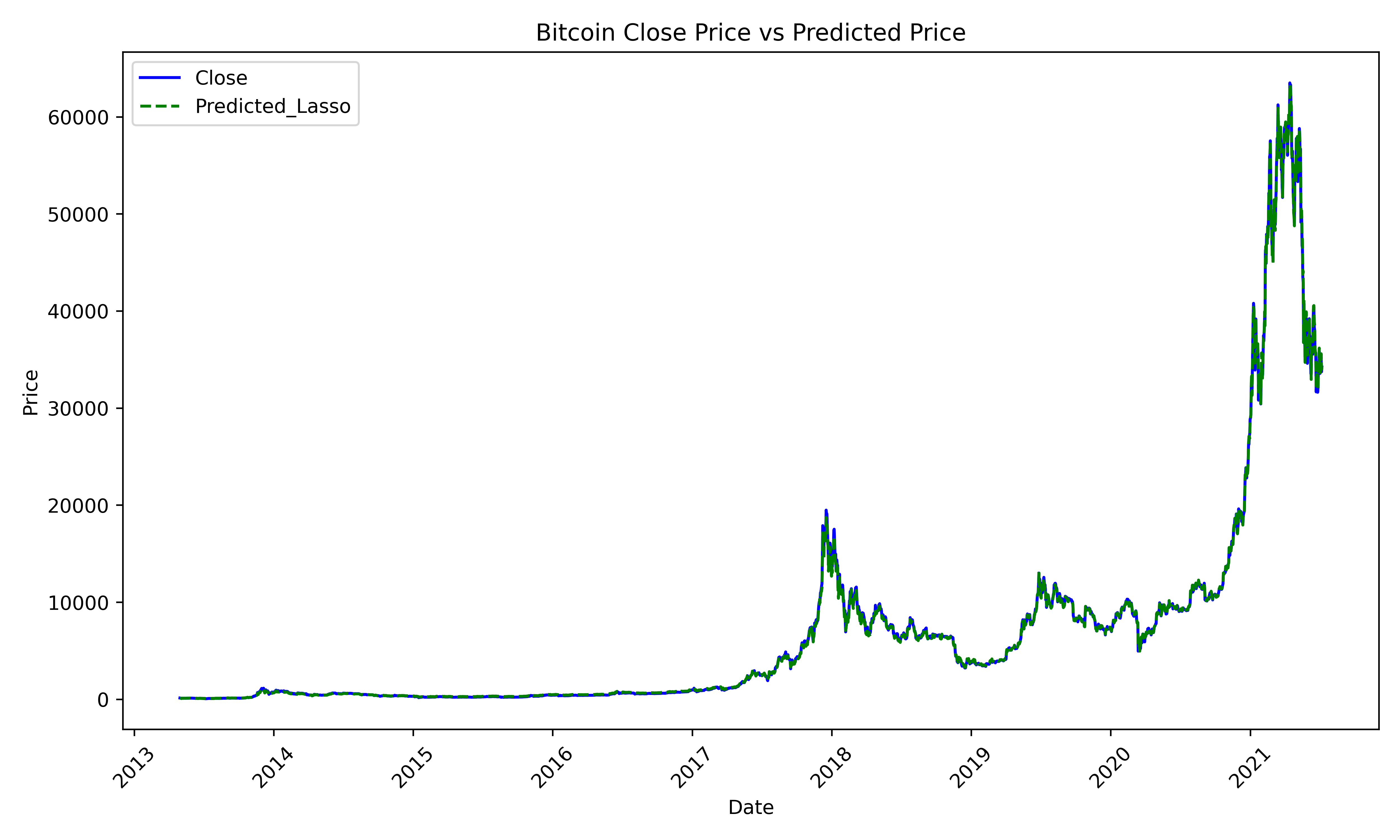}
    \caption{Predicted value versus real value of bitcoin closing prices by configured \textbf{\textit{Linear Regression}}(left) and \textbf{\textit{Lasso}}(right), spanning from 2013-04 to 2021-07, note that \textbf{\textit{Lasso}} outperforms other rivalries in this task.}
    \label{figure-vis}
\end{figure}

Experimental results are displayed in the Table \ref{figure-result}. It's obvious that \textbf{\textit{Lasso Regression}} outperforms other methods in terms of MAE(mean absolute error), further prove our assumption in Section \ref{sec-exp}. \textbf{\textit{Linear Regression}} owns the largest confidence score 0.9998. Predicted results are shown in Figure \ref{figure-vis}.

\begin{table}
 \caption{Results of candidate models in bitcoin prices forecasting. 'MAE' column signifies the mean absolute errors to reflect the grounded performance of every predictor, the lower the better; 'Score' column represents the determination of the prediction, the higher the better. Best predictor on metrics are annotated by \textbf{bold} style.}
  \centering
  \begin{tabular}{lll}
    \toprule
    Method     &  MAE(Val)  & Score   \\
    \midrule
    \textbf{\textit{Naive Linear Regression}} & 96.20 & \textbf{0.9998}  \\
    \textbf{\textit{Lasso Regression}} & \textbf{91.80} & 0.9997 \\
    \textbf{\textit{Decision Tree}} & 135.02 & 0.9988 \\
    \bottomrule
  \end{tabular}
  \label{figure-result}
\end{table}

\section{Discussion}\label{sec-diss}
This section focus on the explanability of conventional machine learning approaches, with a handful of practical advice.

\textbf{Decision Trees.}
Decision trees are a formidable tool for prognosticating Bitcoin time-series closing prices. They proficiently segregate data based on features like historical prices, trading volumes, and external market indicators. This method unfolds recursively, partitioning data with a focus on informative features. The resultant tree structure offers intuitive comprehension of the conditions leading to specific price predictions. A comparative analysis reveals that decision trees excel in scenarios with distinct feature-based segmentation but may struggle with highly nonlinear or complex relationships.

\textbf{Linear Regression.}:
Within the context of Bitcoin closing price prediction, linear regression emerges as a venerable paradigm, where past closing prices assume the role of pivotal input features. This classical approach entails fitting a linear equation to historical data, thereby enabling the anticipation of future closing prices predicated on discerned linear relationships. The straightforward and interpretable nature of linear regression aligns seamlessly with the aim of comprehending and predicting Bitcoin time-series dynamics.

\textbf{Lasso Regression.}:
In the realm of Bitcoin time-series forecasting, Lasso regression, an extension of linear regression incorporating L1 regularization, assumes a pivotal role. This method adeptly mitigates overfitting concerns by penalizing the absolute magnitude of regression coefficients. Lasso regression, with its intrinsic sparsity-inducing mechanism, serves as an indispensable tool for feature selection, particularly efficacious when contending with an abundance of input features.

\section{Conclusion}

\subsection*{Conclusion and Future Directions}

Our research marks a significant contribution to the field of cryptocurrency analysis, presenting a comprehensive framework for Bitcoin price prediction that synergizes the interpretability of statistical analysis with the predictive prowess of advanced machine learning models. Through extensive experimentation, we discovered that the Lasso Regression model outperforms its counterparts – Linear Regression, Convolutional Neural Networks (CNN), and Autoregressive Integrated Moving Average (ARIMA) – in terms of accuracy and reliability.

\subsection*{Key Findings and Implications}

The superior performance of Lasso Regression can be attributed to its ability to handle large datasets and select relevant features, thereby reducing overfitting and enhancing prediction accuracy. This finding is pivotal for investors and financial analysts, as it underscores the importance of feature selection and regularization in forecasting cryptocurrency prices.

\subsection*{Real-World Applications}

Our methodology extends beyond academic interest, offering tangible benefits to various stakeholders in the cryptocurrency market. Investors can leverage our predictive models for strategic portfolio management and risk assessment. Financial institutions may utilize these insights for asset allocation and market trend analysis. Moreover, policymakers can better understand market dynamics, which is crucial for regulatory and economic planning in the evolving landscape of digital currencies.

\subsection*{Potential for Broader Impact}

The implications of our research are not confined to Bitcoin alone. The proposed framework can be adapted and applied to other cryptocurrencies, providing a versatile tool for analyzing a range of digital assets. Additionally, our approach can be integrated with traditional financial models, offering a more holistic view of the financial market.

\subsection*{Future Research Directions}

Looking ahead, we envision several promising directions for future research. Exploring hybrid models that combine statistical methods with deep learning could yield more robust and accurate predictions. Investigating the impact of external factors, such as regulatory changes or macroeconomic indicators, on cryptocurrency prices could provide deeper insights. Furthermore, extending our methodology to real-time data analysis could enhance its practical utility, offering up-to-the-minute forecasts in a fast-paced market.

\subsection*{Concluding Thoughts}

In summary, our research provides a novel and effective approach to Bitcoin price prediction, blending traditional statistical methods with modern machine learning techniques. Its potential applications in investment strategy, financial analysis, and policy formulation highlight its relevance and utility in the rapidly evolving world of cryptocurrencies. As the digital currency landscape continues to grow and evolve, our work lays a foundation for future innovations in cryptocurrency analysis and forecasting.

\section*{Author contributions}
Liu S. conceived the experiments and proposes the research methodology, Liu S. and Wu K. conducted the experiments, Wu K. crafted the literature review, Liu S., Wu K. and Jiang C. analysed the results. All authors reviewed the manuscript and formatted the manuscript.

\label{conclusion}

\bibliographystyle{unsrt}  
\bibliography{references}  
\end{document}